\documentclass[conference]{IEEEtran}
\IEEEoverridecommandlockouts
\usepackage{cite}
\usepackage{amsmath,amssymb,amsfonts}
\usepackage{algorithmic}
\usepackage{graphicx}
\usepackage{textcomp}
\usepackage{xcolor}
\usepackage{listings}
\usepackage{url}

\usepackage{xcolor}

\newcommand\todo[1]{}

\newcommand\anon[1]{#1}

\newcommand\code[1]{\lstinline{#1}}

\definecolor{codegreen}{rgb}{0,0.6,0}
\definecolor{codegray}{rgb}{0.5,0.5,0.5}
\definecolor{codepurple}{rgb}{0.58,0,0.82}
\definecolor{backcolour}{rgb}{1.0, 1.0, 1.0}

\lstdefinestyle{mystyle}{
	backgroundcolor=\color{backcolour},   
	commentstyle=\color{codegreen},
	keywordstyle=\color{magenta},
	numberstyle=\tiny\color{codegray},
	stringstyle=\color{codepurple},
	basicstyle=\ttfamily\footnotesize,
	breakatwhitespace=false,         
	breaklines=true,                 
	xleftmargin=1cm,
	captionpos=b,                    
	keepspaces=true,                 
	numbers=left,                    
	numbersep=5pt,                  
	showspaces=false,                
	showstringspaces=false,
	showtabs=false,                  
	tabsize=1
}

\makeatletter
\def\lst@makecaption{%
	\def\@captype{table}%
	\@makecaption
}
\makeatother

\begin{document}

\title{CaiRL: A High-Performance Reinforcement Learning Environment Toolkit}

\author{\IEEEauthorblockN{Per-Arne Andersen}
	\IEEEauthorblockA{\textit{Department of ICT} \\
		\textit{University of Agder}\\
		Grimstad, Norway \\
		per.andersen@uia.no}
	\and
	
	\IEEEauthorblockN{Morten Goodwin}
	\IEEEauthorblockA{\textit{Department of ICT} \\
		\textit{University of Agder}\\
		Grimstad, Norway \\
		morten.goodwin@uia.no}
	\and
	
	\IEEEauthorblockN{Ole-Christoffer Granmo}
	\IEEEauthorblockA{\textit{Department of ICT} \\
		\textit{University of Agder}\\
		Grimstad, Norway \\
		ole.granmo@uia.no}
	
}


\maketitle

\begin{abstract}
This paper addresses the dire need for a platform that efficiently provides a framework for running reinforcement learning (RL) experiments. We propose the CaiRL Environment Toolkit as an efficient, compatible, and more sustainable alternative for training learning agents and propose methods to develop more efficient environment simulations.

There is an increasing focus on developing sustainable artificial intelligence. However, little effort has been made to improve the efficiency of running environment simulations. The most popular development toolkit for reinforcement learning, OpenAI Gym, is built using Python, a powerful but slow programming language. We propose a toolkit written in C++ with the same flexibility level but works orders of magnitude faster to make up for Python's inefficiency. This would drastically cut climate emissions.	

CaiRL also presents the first reinforcement learning toolkit with a built-in JVM and Flash support for running legacy flash games for reinforcement learning research. We demonstrate the effectiveness of CaiRL in the classic control benchmark, comparing the execution speed to OpenAI Gym. Furthermore, we illustrate that CaiRL can act as a drop-in replacement for OpenAI Gym to leverage significantly faster training speeds because of the reduced environment computation time.
\end{abstract}


\section{Introduction}
\label{sec:introduction}

Reinforcement Learning (RL) is a machine learning area concerned with sequential decision-making in real or simulated environments. RL has a solid theoretical background and shows outstanding capabilities to learn control in unknown non-stationary state-spaces \cite{Berkenkamp2017, cheung20a, Padakandla2020}. Recent literature demonstrates that Deep RL can master complex games such as Go \cite{Schrittwieser2020a}, StarCraft II \cite{Vinyals2019}, and progressively move towards mastering safe autonomous control \cite{Berkenkamp2017}. Furthermore, RL has the potential to contribute to health care for tumor classification \cite{Yu2019}, finances \cite{Li2017}, and industry-4.0 \cite{Pane2019} applications. RL solves problems iteratively by making decisions while learning from received feedback signals.

\subsection{Research Gap}
However, fundamental challenges limit RL from working reliably in real-world systems.  The first issue is that the exploration-exploitation trade-off is difficult to balance in real-world systems because it is also a trade-off between successful learning and safe learning  \cite{Berkenkamp2017}. The reward-is-enough hypothesis suggests that RL algorithms can develop general and multi-attribute intelligence in complex environments by following a well-defined reward signal. However, the second challenge is that reward functions that lead to efficient and safe training in complex environments are difficult to define \cite{Silver2021}. Given that it is feasible to craft an optimal reward function, agents could quickly learn to reach the desired behavior but still require exploration to find a good policy. RL also requires many samples to learn an optimal behavior, making it difficult for a policy to converge without simulated environments. While there are efforts to address RL's safety and sample efficiency concerns, it remains an open question \cite{Moerland2020}. These concerns represent challenges for training RL algorithms climate-efficient in an environmentally responsible manner. Most RL algorithms require substantial calculations to train and simulate environments before achieving satisfactory data to conclude performance measurements. Therefore, current state-of-the-art methods have a significant negative impact on the climate footprint of machine learning \cite{Henderson2020}.

Because of the difficulties mentioned above, a large percentage of RL research is conducted in environment simulations. Learning from a simulation is convenient because it simplifies quantitative research by allowing agents to freely make decisions that learn from catastrophic occurrences without causing harm to humans or real systems. Furthermore, simulations can operate quicker than real-world systems, addressing some of the issues caused by low sample efficiency.

There are substantial efforts in the RL field that focus on improving sample efficiency for algorithms but little work on improving simulation performance through implementation or awareness. Currently, most environments and simulations in RL research are integrated, implemented, or used through the Open AI Gym toolkit. The benefit of using AI Gym is that it provides a common interface that unifies the API for running experiments in different environments. There are other such efforts like Atari 2600 \cite{Bellemare2012}, Malmo Project \cite{Johnson2016a}, Vizdoom \cite{Kempka2016b}, and DeepMind Lab \cite{Beattie2016a}, but there is, to the best of our knowledge, no toolkit that competes with the environment diversity seen in AI Gym. AI Gym is written in Python, an interpreted high-level programming language, leading to a significant performance penalty. At the same time, AI Gym has substantial traction in the RL research community. Our concern is that this gradually leads to more RL environments and problems being implemented in Python. Consequently, RL experiments may cause unnecessary computing costs and computation time, which results in a higher carbon emission footprint \cite{Zhang2022a}. Our concern is further increased by comparing the number of RL environment implementations in Python versus other low-level programming languages. 

Our contribution addresses this gap by developing an alternative to AI Gym without these adverse side effects by offering a comparable interface and increasing computational efficiency. As a result, we hope to reduce the carbon emissions of RL experiments for a more sustainable AI.

\subsection{Contribution Scope}
We propose the CaiRL Environment toolkit to fill the gap of a flexible and high-performance toolkit for running reinforcement learning experiments. CaiRL is a C++ interface to improve setup, development, and execution times. Our toolkit moves a considerable amount of computation to compile time, which substantially reduces load times and the run-time computation requirements for environments implemented in the toolkit. CaiRL aims to have a near-identical interface to AI Gym, ensuring that migrating existing codebases requires minimal effort. As part of the CaiRL toolkit, we present, to the best of our knowledge, the first Adobe Flash compatible RL interface with support for Actionscript 2 and 3. 

Additionally, CaiRL supports environments running in the Java Virtual Machine (JVM), enabling the toolkit to run Java seamlessly if porting code to C++  is impractical. Finally, CaiRL supports the widely used AI Gym toolkit, enabling existing Python environments to run seamlessly. Our contributions summarize as follows:
\begin{enumerate}
	\item Implement a more climate-sustainable and efficient experiment execution toolkit for RL research.
	\item Contribute novel problems for reinforcement learning research as part of the CaiRL ecosystem.
	\item Empirically  demonstrate the performance effectiveness of CaiRL.
	\item Show that our solution effectively reduces the carbon emission footprint when measuring following the metrics in \cite{henderson2020towards}.
	\item Evaluate the training speed of CaiRL and AI Gym and empirically verify that improving environment execution times can substantially reduce the wall-clock time used to learn RL agents.
\end{enumerate}

\subsection{Paper Organization}
In Section 2, we dive into the existing literature on reinforcement learning game design and compare the existing solution to find the gap for our research question. Section 3 details reinforcement learning from the perspective of CaiRL and the problem we aim to solve. Section 4 details the design choices of CaiRL and provides a thorough justification for design choices. Section 5 presents our empirical findings of performance, adoption challenges, and how they are solved, and finally compares the interface of the CaiRL framework to OpenAI Gym (AI Gym). Section 6 presents a brief design recommendation for developers of new environments aimed at reinforcement learning research. Finally, we conclude our work and outline a path forwards for adopting CaiRL.


\section{Background}
\label{sec:background}

\subsection{Reinforcement Learning}
Reinforcement Learning is modeled according to a Markov Decision Process (MDP) described formally by a tuple \((S, A, T, R, \gamma, s_0)\). \(S\) is the state-space, \(A\) is the action-space, \(T \colon S \times A \rightarrow S\) is the transition function, \(R \colon S \times A  \rightarrow \mathbb{R}\) is the reward function~\cite{Sutton2018}, \(\gamma\) is the discount factor, and \(s_0\) is starting state. In the context of RL, the agent operates iteratively until reaching a terminal state, at which time the program terminates. Q-Learning is an off-policy RL algorithm and seeks to find the best action to take given the current state. The algorithm operates off a Q-table, an n-dimensional matrix that follows the shape of state dimensions where the final dimension is the Q-values. Q-Values quantify how good it is to act \(a\) at time \(t\). This work uses Deep Learning function approximators in place of Q-tables to allow training in high-dimension domains \cite{Mnih2015}. This forms the algorithm Deep Q-Network (DQN), one of the first deep learning-based approaches to RL, and is commonly known for solving Atari 2600 with superhuman performance \cite{Mnih2015}. Section \ref{sec:carbon_emission} demonstrates that our toolkit significantly reduces the run-time and carbon emission footprint when training DQN in traditional control environments.

\subsection{Graphics Acceleration}
\label{sec:graphics-accel}
\todo{Introduction}
A graphics accelerator or a graphical processing unit (GPU) intends to execute machine code to produce images stored in a frame buffer. The machine code instructions are generated using a rendering unit that communicates with the central processing unit (CPU) or the GPU. These methods are called software rendering or hardware rendering, respectively. GPUs are specialized electronics for calculating graphics with vastly superior parallelization capabilities to their software counterpart, the CPU. Therefore, hardware rendering is typically preferred for computationally heavy rendering workloads. Consequently, it is reasonable to infer that hardware-accelerated graphics provide the best performance due to their improved capacity to generate frames quickly. On the other hand, we note that when the rendering process is relatively basic (e.g., 2D graphics) and access to the frame buffer is desired, the expense of moving the frame buffer from GPU memory to CPU memory dramatically outweighs the benefits. \cite{mileff2012efficient}

\todo{About, and why it is central to CaiRL?}
According to \cite{mileff2012efficient}, software rendering in modern CPU chips performs 2-10x faster due to specialized bytecode instructions. This study concludes that the GPU can render frames faster, provided that the frame permanently resides in GPU memory. Having frames in the GPU memory is impractical for machine learning applications because of the copy between the CPU and GPU. The authors in \cite{Mendel1348908} propose using Single Instruction Multiple Data (SIMD) optimizations to improve game performance. SIMD extends the CPU instruction set for vectorized arithmetic to increase instruction throughput. The authors find that using SIMD instructions increases performance by over 80\% compared to traditional CPU rendering techniques. 

The findings in these studies suggest that software acceleration is beneficial in some graphic applications, and similarly, we find it useful in a reinforcement learning context. Empirically, software rendering performs better for simple 2D and 3D graphic applications due to the high-latency copy operation needed between the GPU and CPU. Much of the success of CaiRL lies in the fact that software rendering, while being slower for advanced games such as StarCraft, significantly outperforms hardware rendering for simple graphics. One alternative to improve performance in hardware rendering is to use pixel buffer objects or an equivalent implementation. A pixel buffer object (PBO) is a buffer storage that allows the user to retrieve frame buffer pixels asynchronously while a new frame buffer is drawn to the screen frame buffer. In particular, copying pixels without PBO is slow because rendering must halt while the buffer is read  \cite{Lawlor2009}. 

\subsection{Programming Languages}
\label{sec:programming_languages}
\todo{Introduction}

Machine learning research and application development have been carried out in various programming languages throughout history. In more recent history, the Python language has been used more frequently in the scientific community and, more specifically, in machine learning, and deep learning \cite{Raschka2020}. Unfortunately, Python's most used implementation is CPython, a single-threaded implementation with little regard for efficiency compared to compiled languages. However, Python's most popular toolkits for machine learning are implemented in compiled languages and use wrapper code to interact to increase performance. A study by Zehra et al. suggests that C++ has approximately a 50 times performance advantage over Python, and Python has advantages in code readability for beginners in programming \cite{Zehra2020}. It is clear from these studies that Python is great for prototyping and learning programming but is not suitable for performance-sensitive tasks. It is natural to seek an approach that can preserve the simplicity of Python while also maintaining acceptable task execution performance.

Pybind11 is one such framework that provides a method to create an efficient bridge between C++ and Python code. Pybind11 is a lightweight library that exposes C++ types in Python and vice versa but focuses mainly on exposing C++ code paths to Python applications. There is a minor performance penalty during the conversion between Python and C++ objects. Hence, implementations in C++ will run at near-native performance in Python. For this reason, we follow the path of implementing an efficient experiment toolkit for reinforcement learning in C++ with binding code to allow Python to interface with CaiRL.

\subsection{Summary}
The goal of CaiRL is to create an expanding set of high-performance environments for RL research. It is essential to encourage good practices by adding novel environments to the toolkit. Our extensive practical testing finds that rendering graphics in software provides substantially higher throughput for applications where access to the frame buffer is desirable. This observation is especially prominent for simple 2D and 3D-based applications. However, the benefits diminish as the graphical complexity increases. For example, it is clear from our findings that games such as StarCraft II render better using hardware acceleration. 

We study the implications of implementation language for CaiRL and find that the choice of programming language is essential to CaiRL because it aims to be efficient and reduce the carbon emission footprint as much as possible. C++ seems like a natural choice as it is mature, has a stable standard library, and supersets the C language. 

\section{Design Specifications}
\label{sec:design_specs}
The design goal of CaiRL is to have interoperability with AI Gym, but with orders of magnitude better performance and flexibility to support environments in a multitude of programming languages. Keeping full compatibility with AI Gym is central to trivializing the two frameworks without significant amendments to existing code. 

CaiRL is a novel reinforcement learning environment toolkit for high-performance experiments. By designing such a toolkit, reinforcement learning becomes more affordable due to reduced execution costs and strives toward more sustainable AI. A bi-effect of these goals is that experiments run significantly faster, and most CPU cycles are spent on training AI instead of evaluating game states. The CaiRL environment toolkit supports classical RL problems such as (1) Cart-Pole, Acro-Bot, Mountain-Car, and Pendulum, (2) Novel, high-complexity games such as \anon{Deep RTS, \cite{Andersen2018a}, Deep Line Wars, X1337 Space Shooter}, and (3) over 1 000 flash games available for experimentation. \footnote{We invite the reader to \anon{\url{http://github.com/cair/rl}} for detailed toolkit documentation.} 

The engine of CaiRL relies upon C++ with highly performant fast-paths such as Single Instruction Multiple Data (SIMD) for vectorized calculation that fits into the processor registry in a single instruction. The design of CaiRL mimics AI Gym but relies on templating and \code{const} expressions to evaluate calculations at compile-time instead of run-time. CaiRL is split into modules, and we dedicate this section to describing the design decisions and the resulting interaction layer and benefits compared to similar solutions.

\subsection{Building Blocks}
CaiRL follows the module design pattern to have minimal cross-dependencies between toolkit components. This has several benefits, namely (1) being easier to maintain and (2) reducing compile times significantly. CaiRL is composed of six essential modules:
\begin{enumerate}
	\item \code{Runners} is a bridge for accessing non-native run-times, enabling a unified API for all environments. Flash environments use the Lightspark runner to run Flash games seamlessly. Similarly, Java games have a specialized Java Virtual Machine (JVM) runner.
	
	\item \code{Renderers} is a module for drawing graphical frame buffer output to the screen. Currently, Blend2D and OpenCV are part of this module. This module is essential for training agents in graphical environments.
	
	\item \code{Environments} are the module for integrating games and applications with a unified interface. This interface is near-identical to AI Gym but has less overhead because of the more efficient precompilation of machine code.

	\item \code{Wrappers} are also similar to what is found in AI Gym. This module features code to wrap environment instances to change the execution behavior, such as limiting the number of timesteps before reaching the terminal state. The initial version of CaiRL features wrappers to flatten the state observation and add max timestamp restrictions.
	
	\item \code{Spaces} are a module for defining the shape of state observation and action spaces, similar to AI Gym. All of the spaces use highly optimized code, which efficiently increases populating data matrices. The Box type features n-dimensional matrices, and finally, the Discrete type defines a one-dimensional vector of integers.
	
	\item \code{Tooling} is the module for contributions that reach a stable state and enrich the features of CaiRL. One such example is the tournament framework that trivializes running single-elimination and Swiss-based tournaments.
\end{enumerate}
The CaiRL toolkit has exposed interfaces through its native C++ API and the Python API. The binding code is automatically generated for environments following the standard definition found in the \code{Env} class, but for highly customized implementations, such bindings must be added manually. Similarly, the CaiRL toolkit compiles Python-compatible machine code with significantly lower overhead when loaded and interpreted by CPython. See the discussion in Section \ref{sec:programming_languages}.

\subsection{Implementation Layer}
There are two ways of building reinforcement learning environments with CaiRL (1) using C++ directly or (2) through the Python to C++ bindings. CaiRL performs well in Python and C++ because most of the computation runs in optimized code. However, the Python bindings have additional computational costs because each line is interpreted and translated from Python and C++. The interpreter overhead can be reduced by diverging from the normal AI Gym API and implementing a \code{run} function, notably eliminating the need for interpreted loop code in Python. The primary goal of the CaiRL API is to match the AI Gym API to enable a seamless experience when migrating existing codebases to CaiRL.
\begin{lstlisting}[language=C++,label={lst:minicairl}, caption=Minimal Example of CaiRL-CartPole-v1 in C++]
	e =  Flatten<TimeLimit<200,CartPoleEnv>>()
	for(int ep = 0; ep < 100; ep++){
		e.reset();
		int term, steps = 0;
		while(!term){
			steps++;
			const auto [s1, r, term, info] = 
			e.step(e.action_space.sample());
			auto obs = e.render();
		}
	}
\end{lstlisting}

Listing \ref{lst:minicairl} shows the C++ interface of the CaiRL toolkit. In C++, we deviate from the AI Gym API to allow modules as static template classes, as seen in line 1. A template defines a class that evaluates much of the program logic during compile-time. This has considerable run-time benefits because code initialization is done during compile-time. The downsides are that compile times increase substantially, and polymorphism is impossible between Python classes and C++ templates. However, it is possible to alleviate these challenges by predefining classes from the template implementations. This allows contributors to add Python-based environments to the repository of available experiments, however, at the cost of providing diminishing performance benefits.

A very central component of CaiRL is the ability to run experiments natively in Python. This becomes possible by creating code that interfaces C++ and Python using Pybind11. Pybind11 is a library that provides the ability to call code from the CaiRL shared library (C++ machine code) and the Python interpreter efficiently. There is no need for C++ experience using the Python binding code, and it is possible to use and customize CaiRL with Python for specialized experiments. The Python interface is similar to the C++ interface but focuses more on compatibility with the AI Gym interface.

\begin{lstlisting}[language=python,label={lst:minicairpython}, caption=Minimal Example of AI Gym and CaiRL CartPole-v1 in Python]
	#e =  gym.make("CartPole-v1")
	e =  cairl.make("CartPole-v1") # Use CaiRL
	for ep in range(100):
		e.reset()
		term, steps = 0
		while not term:
			steps++
			a = e.action_space.sample()
			s1, r, term, info = e.step(a)
			obs = e.render()
\end{lstlisting}
Listing \ref{lst:minicairpython} illustrates the use of CaiRL in Python compared to AI Gym. In particular, to change between AI Gym and CaiRL, the only change required is to use the cairl package (Line 2) instead of the gym package (Line 1).

\subsection{Affordable and Sustainable AI}
AI is a constantly growing field of research, and with the shifted focus on Deep Learning, it is well understood that the need for computing power has increased sharply. Deep Learning models have a range of a few thousand parameters, up to several billion parameters that require carefully tuning with algorithms such as stochastic gradient descent. Hence, compute power plays an essential role in the performance of the trained model. The same applies in Deep RL but requires extensive data sampling from an environment. It is reasonable to conclude that the cost of conducting trials increases rapidly and contributes against the emergence of more sustainable AI. CaiRL aims to minimize the cost of reinforcement learning by reducing environment execution time. In essence, this has the bi-effect of reducing the carbon emission footprint in RL significantly compared to existing solutions, as observed in section \ref{sec:perf_eval}.

\section{Game Run-times and  Platforms}
This section presents the primary run-times that CaiRL supports to integrate environments from run-times other than Python and C++ seamlessly.  

\subsection{JVM Applications}
Java is a popular programming language that runs in the Java Virtual Machine (JVM). Although Java is not the dominant language for environments in the RL research community, there are a few notable examples, such as MicroRTS \cite{Ontanon2013} and the Showdown AI competition \cite{Lee2017a}. These environments have shown significant value to several research communities in reinforcement learning, evolutionary algorithms, and planning-based AI. To integrate JVM-based games in CaiRL, the programmer defines configuration in a CMake file that describes how the source code is built to a Java archive (JAR) file. Then the programmer defines a C++ class that extends the \code{Env} class interface. The JVM and the C++ machine code communication is through the Java Native Interface (JNI). Using the JNI bridge, it is trivial to create a mapping from C++ to JVM, and it is conveniently also performant as the JVM has good optimization options. There are similar efforts to bridge Java games through JNI for games such as MicroRTS  \cite{Huang2021}, but CaiRL aims toward a generic approach that encapsulates many existing games.

\subsection{Python Environments}
Python is arguably the most used programming language for RL research in recent literature, as suggested by Github tag statistics. We perform the following the search queries: \code{topic:reinforcement-learning+topic:game+language:python} for finding relevant Python environments, and \code{topic:reinforcement-learning+topic:game+language:c++} for C++ environments. We observe a ratio of 114:1 in favor of Python. Consequently, many of the popular reinforcement learning environments have native Python implementations. We approach the task of improving such environments with two possible solutions. The first approach automatically converts Python code into C++ using the Nuitka library found at \url{https://github.com/Nuitka/Nuitka}. It is also possible to add environments directly as a CaiRL Python package, although this method does not improve the performance and does not address climate emission concerns. All third-party environments reside in the \code{cairl.contrib} package and are freely available through the C++ and Python interface. For an environment to be fully compatible with the CaiRL interface, the environment must inherit the abstract \code{Env} class and implement the \code{step(action)}, \code{reset()} , and \code{render()} function. However, there are several open questions on how to efficiently improve the performance of most environments implemented in Python, see Section \ref{sec:future_work}.

\subsection{Flash Run-time}
The most notable feature of CaiRL is the ability to run flash games without external applications. CaiRL extends the LightSpark flash emulator for Actionscript 3 and falls back to GNU Gnash for ActionScript 2. CaiRL features a repository of over 1300 flash games for conducting AI research and reinforcement learning research. In this paper, we focus on the Multitask environment. Multitask is an environment that provides minigames that the agent must control concurrently. If the agent fails one of the tasks, the game terminates. The reward function is defined as positive rewards while the game is running and negative rewards when the game engine terminates (e.g., end of  the game), indicating that the game is lost. The game observations are either raw pixels or the virtual Flash memory, and the actions-space is discrete. Our observation is that most existing Flash games have short-horizon episodes with few objectives to reach a positive terminal state. In addition, many of the games have simple game rules that are especially suited for benchmarking non-hierarchical RL algorithms. The CaiRL flash runner substantially expands the number of available game environments for experiments. To the best of our knowledge, CaiRL is the only tool that can control the game execution speed and guarantee broad support for Actionscript 2 and 3.

\subsection{Puzzle Run-time}
CaiRL supports the comprehensive collection of puzzles from the Simon Tatham collection\cite{Bauer2021}. This collection aims to provide logical puzzles that are solvable either by humans or algorithms. While reinforcement learning is not mainly known for solving logical puzzles, some literature suggests that RL can solve puzzles \cite{Dandurand2012}, potentially with the options framework from \cite{Sutton1999}. We find it beneficial to add puzzles for future research and demonstrate flexibility in adding new problems and environments. All puzzles include a heuristic-based solver, enabling transfer and curriculum learning research.

\section{Evaluations}

\subsection{Performance Evaluation}
\label{sec:perf_eval}
To evaluate the performance of CaiRL, we compare the classic control environments from AI Gym with an identical implementation using the CaiRL toolkit. Experiments run for 100 000 timesteps, and the measurements are averaged over 100 consecutive trials. The environments are evaluated with and without graphical rendering to demonstrate the effectiveness of raw computation speed and software rendering. 

\begin{figure}[ht]
	\includegraphics[width=\linewidth]{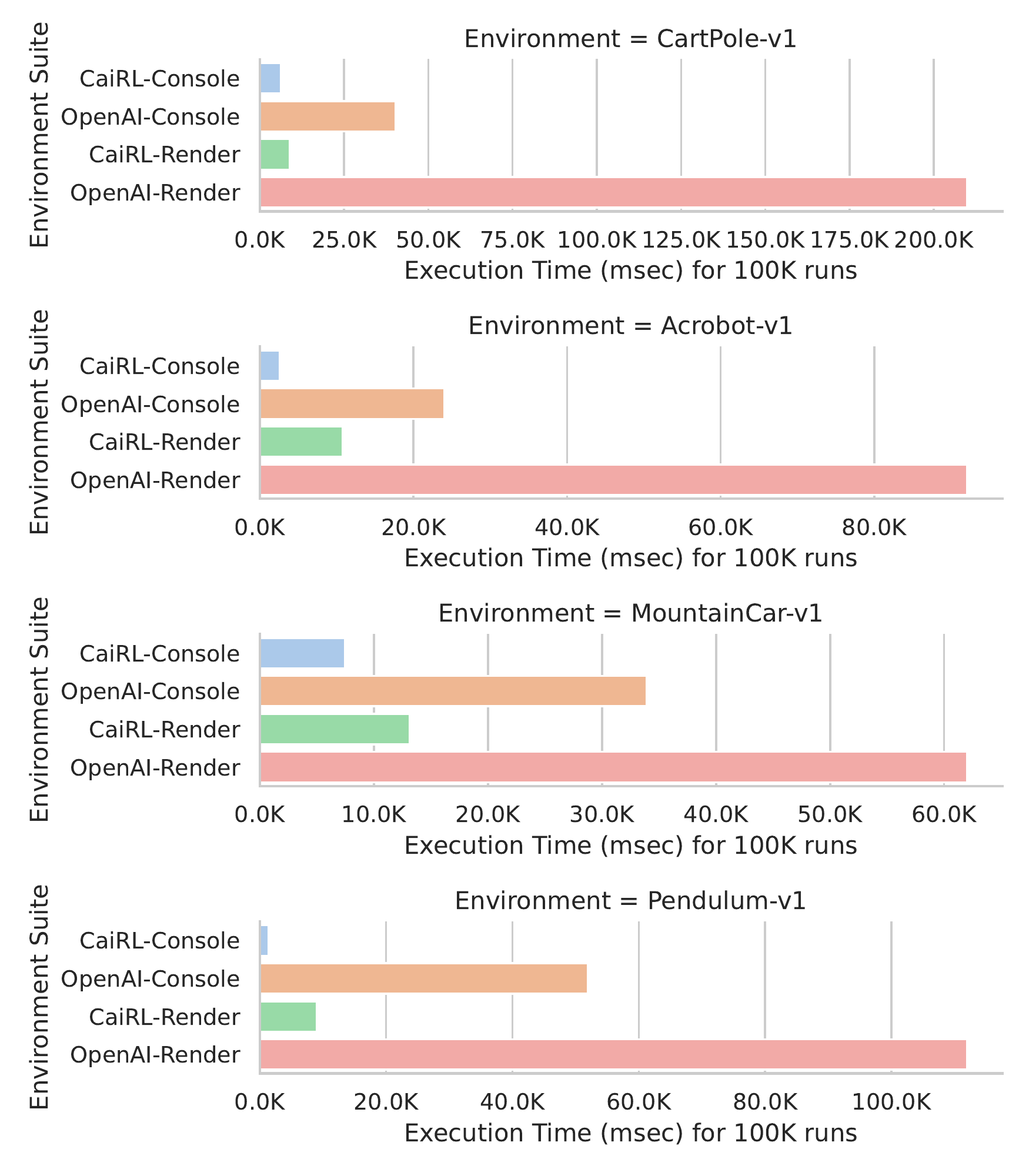}
	\label{fig:demo_0}
	\caption{Execution run-time evaluation between CaiRL and AI Gym in the classical control tasks. The x-axis illustrates the execution time for 100 000 runs (episodes), averaged over 100 trials. The rows show the console and render versions in CaiRL and AI Gym.}
\end{figure}

Figure \ref{fig:demo_0} demonstrates the average console and rendering performance and clearly shows that CaiRL performs 5x faster in simulations and over 80x faster on rendering than the AI Gym equivalent. The console experiment indicates the raw performance boost when using high-performance programming languages. As discussed in Section \ref{sec:graphics-accel}, the graphical experiment validates the effectiveness of rendering the frame buffer in software instead of using hardware methods. Specifically, the rendering backend in AI Gym utilizes OpenGL and has far greater computational costs when accessing frames, often desirable in RL research.

\subsection{Algorithm Evaluation}
The scope of the algorithm evaluation is two-fold. First, we aim to find if CaiRL implementations can improve training time in that they are measurable, hence having positive effects on economics and climate emission rates. Finally, we evaluate if DQN can improve its behavior using the Flash Run-time in the Multitask game environment. We use the default hyperparameters proposed by \cite{Mnih2015} and use raw images as input to the algorithm for both experiments. The experiments run using an Intel 8700K CPU and an Nvidia GeForce 2080TI.

\begin{figure}[ht]
	\includegraphics[width=\linewidth]{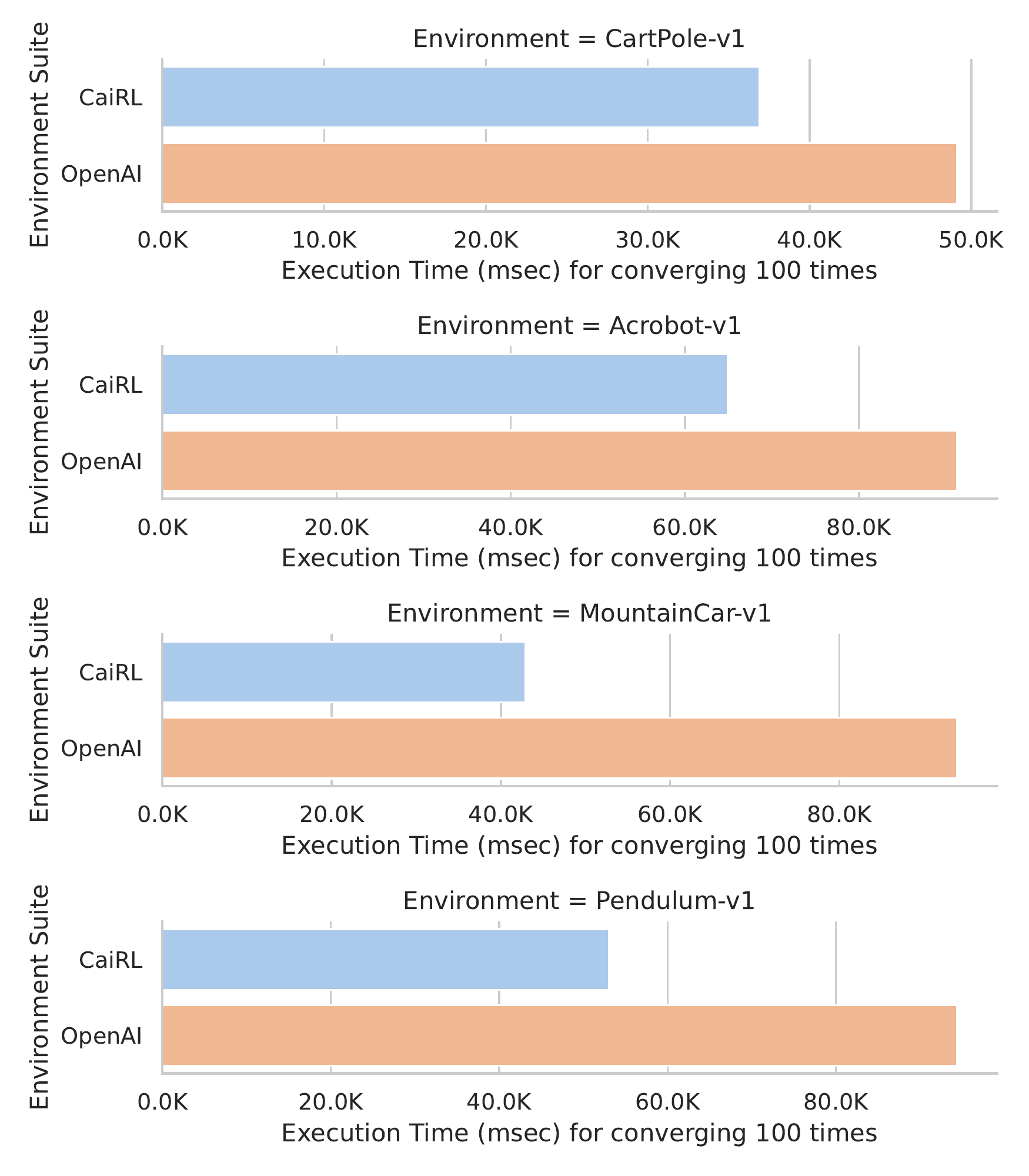}
	\label{fig:classical_control_dqn}
	\caption{The average DQN training time for 100 runs in the classical control environments. The x-axis is the total execution time in milliseconds for training the agent 100 times until reaching the optimal strategy. Each time the agent converges, the policy is reset with a fixed randomization seed.}
\end{figure}

Figure \ref{fig:classical_control_dqn} clearly shows that the DQN algorithm is trained magnitudes faster in the CaiRL environment, indicating that a large part of the training time is the result computation time during sampling the environment. The algorithm trains until mastering the task (stopping criteria) for 100 trials, after which we average the results. Our findings conclusively show substantial wall-clock time reductions for training in the CaiRL environments compared to the AI Gym environments. The average reduction in training time across all trials is roughly 30 percent, illustrating and confirming that efficient environments are essential for developing AI that trains more climate-friendly.

\begin{figure}[ht]
	\includegraphics[width=\linewidth]{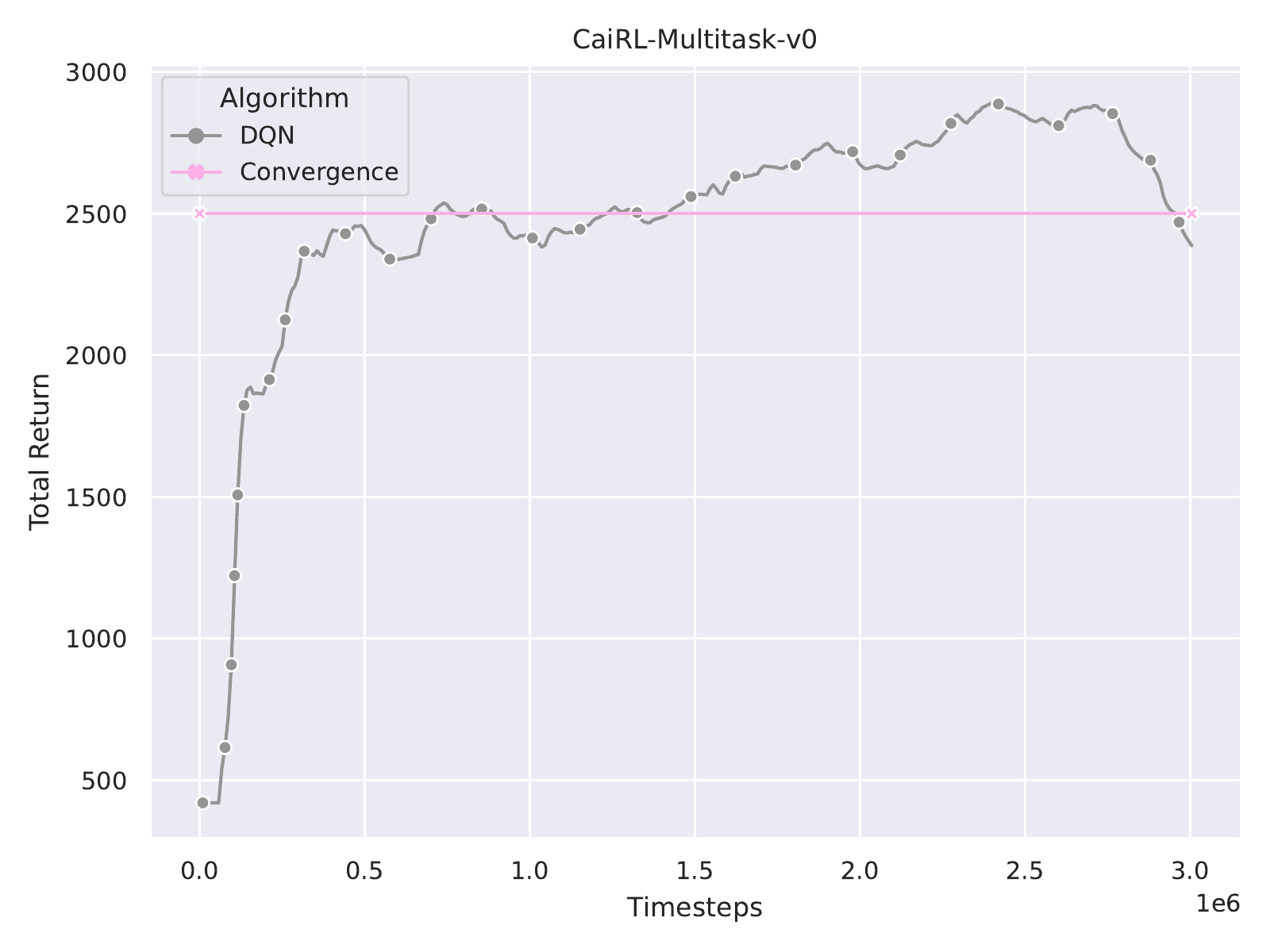}
	\label{fig:multitask_dqn}
	\caption{DQN performance in the Multitask environment. The algorithm solves the environment after approximately 3 000 000 timesteps where the training procedure is averaged over 10 trials.}
\end{figure}

Figure \ref{fig:multitask_dqn} shows that the DQN algorithm successfully solves the multitask environment after approximately 1 500 000 frames averaged over ten trials.  Note that we here only wish to verify that algorithms can learn from the flash game engine. By unlocking the frame rate of the simulation, it is possible to achieve approximately 140 frames per second using Intel 8700K in the Multitask environment. Compared to running the simulation in the integrated flash run-time in browsers, our approach increases the game execution speed to a factor of  4.6x in a majority of flash games. This is because Flash games have the game loop inside the rendering loop. Each training trial took approximately 6 hours to finish, and in total, the experiment lasted for 60 hours.

\subsection{Carbon Emission Evaluation}
\label{sec:carbon_emission}
This section aims to answer the following question:  \textit{Is CaiRL a better alternative for lowering carbon emissions in RL}. To begin answering this question, we rerun experiments with the novel experiment-impact-tracker from \cite{henderson2020towards}. The experiment-impact-tracker is a drop-in method to track energy usage, carbon emissions, and compute utilization of the system and is recently proposed to encourage the researcher to create more sustainable AI. Our experiments run a DQN agent on the classical control environment CartPole-v1 in CaiRL and AI Gym. We compare the toolkits using the console-only version and the graphical variant. We use the following environment configurations and DQN parameters:
\begin{table}[!ht]
	\centering
	\caption{The DQN hyperparameters for the carbon emission experiment}
	\label{tab:dqn_hyperparameters}
	\begin{tabular}{l|l}
		\textbf{Hyperparameter} & \textbf{Value} \\ \hline
		Discount & 0.99 \\
		Units & 32, 32 \\
		Activation & elu \\
		Optimizer & Adam \\
		Loss Function & Huber \\
		Batch Size & 32 \\
		Learning Rate & 3e-4 \\
		Target Update Freq & 150 \\
		Memory Size & 50 000 \\
		Exploration Start & 1.0 \\
		Exploration Final & 0.01
	\end{tabular}
\end{table}

The experiment runs for 1 000 000 timesteps in the console version and 10 000 timesteps for the graphical version\footnote{The experiments code be accessed at \anon{\url{https://github.com/cair/rl}}. }

\begin{table}[!ht]
	\centering
	\caption{The table describes the total carbon emission values and power consumption used during the experiments. The carbon emission is measured in CO2/kg, and the power draw is measured in milliwatt-hour (mWh).}
	\label{tab:co2_results}
	\begin{tabular}{llllll}
		\hline
		Measurement &  Environment &  CaiRL &  Gym &  Ratio \\
		CO2/kg &  Console &  \textbf{0.000014 }&  0.000067 &  20.8955 \\
		CO2/kg &  Graphical & \textbf{0.000051} & 0.075265 &    147578.431373 \\
		Power (mWh) &  Console &  \textbf{0.000319} & 0.001483 & 21.5104 \\
		Power (mWh) &  Graphical & \textbf{0.001131} & 1.673959 & 148006.9849 \\
		\hline
	\end{tabular}

\end{table}

Table \ref{tab:co2_results} shows that CaiRL has a considerably lower carbon emission than AI Gym. CaiRL has 20.89x less carbon emission in the console variant than Gym. The graphical experiment shows a more significant difference with a 147578x reduction in carbon emissions. The reason AI Gym has high emission rates is that it is locked to capturing images from the game window. We measure the emissions by subtracting the DQN time usage with the total time to only account for the environment run-time costs.

\section{ Conclusion}

CaiRL is a novel platform for RL and AI research and aims to reduce program execution time for experiments to reduce budget costs and the carbon emission footprint of AI. CaiRL outperforms AI Gym implementations significantly while also being compatible with existing AI Gym experiments. However, for CaiRL to be effective, code needs to be ported to the CaiRL toolkit. While this may seem tedious, it reduces execution times, reducing RL experiments' economic and climate-related footprint. However, there are preliminary options for automatically translating code from Python to C++, such as using the Nuitka compiler. 

This contribution clearly outlines new recommendations and considerations for developing new environments for RL research. First, we recommend using low-level languages such as C++ to implement the logic and, optionally, using code binding libraries for interoperability between run-times. The effectiveness of this approach is further demonstrated by \cite{Bargiacchi2021, Fua2020}. Second, for 2D graphics, it is clear from our literature review that using software rendering with SIMD capabilities may provide significant benefits when accessing the frame buffer. Following these recommendations, we show that CaiRL has 30\% less overhead than AI Gym.

Lastly, we have illustrated that CaiRL supports many programming languages, including C++, Java, Python, and ActionScript 2 and 3. CaiRL supports over 1300 games in ActionScript, Several C++ games, MicroRTS, and Showdown in Java and supports building python games out of the box. In the evaluations of CaiRL, we demonstrate superiority in performance and positively impact the carbon footprint of AI.

\section{Future Work}
\label{sec:future_work}
This paper has presented CaIRL, a reinforcement learning toolkit for running a wide range of environments from different run-times in a unified framework.

CaiRL is an ambitious project to improve the tools required to conduct efficient reinforcement learning research. In fulfilling its role, the complexity of the toolkit demands extensive testing and verification to ensure that all experiments are performed following the original version to provide reliable experiment results. While CaiRL is now released, several interesting problems potentially can improve the environment performance further. For the continuation of this project, we believe that the following concerns may prove valuable to address:
\begin{itemize}
	\item Find a suitable method of automatic conversion of Python code. Alternatively, be able to run Python code in more efficient run-times, such as the JVM.
	\item Improve the JVM and Flash support so that researchers can more easily add new environments.
	\item Expand the number of run-times that CaiRL supports while maintaining portability and efficiency
	\item Perform static code analysis and recommend code quality improvements and efficiency to further reduce the climate footprint of environments.
\end{itemize}

\bibliographystyle{IEEEtran}

\end{document}